\documentclass{article}
\usepackage[preprint]{neurips_2020}
\usepackage[utf8]{inputenc} 
\usepackage[T1]{fontenc}    
\usepackage{hyperref}       
\usepackage{url}            
\usepackage{booktabs}       
\usepackage{amsfonts}       
\usepackage{nicefrac}       
\usepackage{microtype}      
\usepackage{graphicx,subfigure}
\usepackage{amsmath}
\usepackage{multirow}
\title{LiftPool: Lifting-based Graph Pooling for Hierarchical Graph Representation Learning }

\author{
  Mingxing Xu\\
  Department of Electronic Engineering\\
  Shanghai Jiao Tong University\\
  \texttt{xumingxing@sjtu.edu.cn}\\
  \And
  Wenrui Dai\\
  Department of Computer Science and Engineering\\
  Shanghai Jiao Tong University\\
  \texttt{daiwenrui@sjtu.edu.cn}\\
  \And 
  Chenglin Li\\
  Department of Electronic Engineering\\
  Shanghai Jiao Tong University\\
  \texttt{lcl1985@sjtu.edu.cn}\\
  \And 
  Junni Zou\\
  Department of Computer Science and Engineering\\
  Shanghai Jiao Tong University\\
  \texttt{zoujunni@sjtu.edu.cn}\\
  \And Hongkai Xiong\\
  Department of Electronic Engineering\\
  Shanghai Jiao Tong University\\
  \texttt{xionghongkai@sjtu.edu.cn} \\
}

\newtheorem{pro}{\bf Proposition}

\newtheorem{proof}{\bf Proof}

\begin{document}

\maketitle

\begin{abstract}
Graph pooling has been increasingly considered for graph neural networks (GNNs) to facilitate hierarchical graph representation learning. Existing graph pooling methods commonly consist of two stages, \emph{i.e.}, selecting the top-ranked nodes and removing the rest nodes to construct a coarsened graph representation. 
However, local structural information of the removed nodes would be inevitably dropped in these methods, due to the inherent coupling of nodes (location) and their features (signals). 
In this paper, we propose an enhanced three-stage method via lifting, named LiftPool, to improve hierarchical graph representation by maximally preserving the local structural information in graph pooling. LiftPool introduces an additional stage of graph lifting before graph coarsening to preserve the local information of the removed nodes and decouple the processes of node removing and feature reduction. 
Specifically, for each node to be removed, its local information is obtained by subtracting the global information aggregated from its neighboring preserved nodes. Subsequently, this local information is aligned and propagated to the preserved nodes to alleviate information loss in graph coarsening. 
Furthermore, we demonstrate that the proposed LiftPool is localized and permutation-invariant. The proposed graph lifting structure is general to be integrated with existing down-sampling based graph pooling methods. 
Evaluations on benchmark graph datasets show that LiftPool substantially outperforms the state-of-the-art graph pooling methods in the task of graph classification.  
\end{abstract}

\section{Introduction}
Convolutional neural networks (CNNs) have achieved great success in a variety of challenging tasks, especially in the fields of computer vision and natural language processing \cite{bahdanau2014neural, he2016deep, hinton2012deep, karpathy2014large, krizhevsky2012imagenet},  which is largely owing to the efficient hierarchical representation learning ability of convolution and pooling operations. However, these operations are naturally defined on regular grids with inherent spatial locality and order information, which thus cannot be directly utilized to process non-Euclidean data residing on irregular grids. 
As a matter of fact, graphs can be used to model a large amount of non-Euclidean data, such as biological networks \cite{davidson2002genomic}, social networks \cite{lazer2009social} and chemical molecules \cite{duvenaud2015convolutional}. In recent years, there have been a surge of interest in developing graph neural networks (GNNs) for representation learning over non-Euclidean data by generalizing classical convolution and pooling operations to graph domains \cite{kipf2016semi, velivckovic2017graph, xie2018crystal, zhang2018end}. In this paper, we focus on designing a graph pooling operation that enables hierarchical graph representation learning. 

Graph pooling operation plays an essential and indispensable role since it not only enables GNNs to learn hierarchical graph representations, but also helps to reduce the size of feature maps and parameters, which thus improves learning efficiency and avoids overfitting. Though important, it is a challenging task to generalize classical pooling operations to graphs that are highly irregular and lack the natural notion of locality as well as order information. 
Recently, there are some attempts that can roughly be categorized into two groups: clustering-based  \cite{ying2018hierarchical} and downsampling-based methods \cite{gao2019graph, lee2019self}. The clustering-based methods group nodes with the learned/predefined cluster assignment matrix and construct a coarsened graph with the clustered centroids. Though node features information can be well preserved through feature aggregation, the original graph structures are destroyed. Moreover, these methods suffer from adopting additional networks to learn a dense cluster assignment matrix, whose computational and storage complexity is very heavy (i.e., quadratic to the graph size). Therefore, they cannot scale to large graphs. In contrast, the downsampling-based methods are more efficient by preserving the key features of input graphs. They typically follow a two-stage strategy: nodes to be preserved are firstly selected according to their importance scores that are either calculated with a predefined measure \cite{zhang2018end} or learned with an additional layer \cite{gao2019graph, lee2019self, vinyals2015order}, and then coarsened graphs are constructed by simply removing the rest nodes and their associated features. This process is quiet different from the classical pooing in CNNs where local information can be well preserved by computing a summary for a group of grids with 
an aggregation function(e.g, mean and max). In fact, the nodes (locations) and signals (features) are inherently coupled in graphs, thus the two-stage pooling that simply removes nodes will inevitably result in the information loss of their coupled features that encode their local structural information and node attributes. This local information loss problem not only limits the hierarchical representation learning efficiency and model capacity, but also causes a waste of computational resources.
  \begin{figure}
  \centering
  \includegraphics[width=0.9\columnwidth,height=0.4\columnwidth]{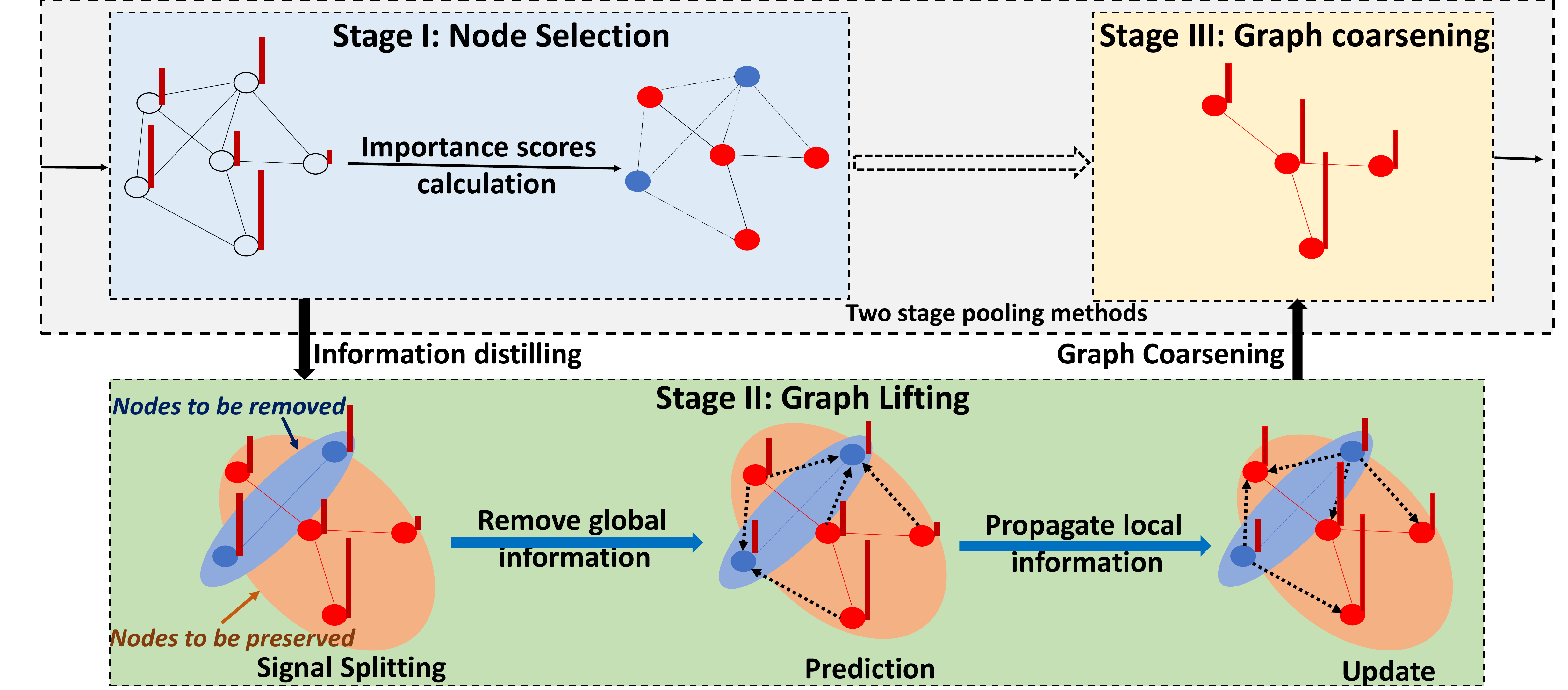}
  \caption{Illustrative diagram of the proposed LiftPool. The upper figure shows previous two-stage pooling methods, consisting of node selection and graph coarsening. An additional graph lifting stage (bottom figure) is introduced by LiftPool: 1) nodes and their associated features are firstly \textit{split} into two subsets; 2) for nodes to be removed, \textit{prediction} operation is then adopted to compute their local information by subtracting global information that can be predicted from the preserved nodes; and 3) finally, the local information of the removed nodes is aligned and propagated into the preserved nodes via \textit{update} operation.}\label{fig1}
  \end{figure}
  
To address the aforementioned information loss problem and better utilize the node attributes information, we propose an enhanced three-stage graph pooling, named LiftPool, which inherits the flexibility and efficiency of downsampling-based pooling methods, while improving the hierarchical graph representation by maximally preserving the local structural information with an additional graph lifting stage. The proposed graph lifting stage decouples the processes of node selection and feature reduction, and is able to propagate local structural information of the removed nodes to the preserved nodes with a novel graph lifting structure before the graph coarsening. Specifically, as illustrated in Fig.~\ref{fig1}, an additional graph lifting stage is introduced by LiftPool. For nodes to be removed, its local structural information is obtained by subtracting the global information aggregated from its neighboring preserved nodes. Subsequently, this local information is aligned and propagated to the preserved nodes to alleviate information loss in graph coarsening. In fact, the lifting process can also be viewed as an information distilling process where local structural information are distilled and concentrated on the preserved nodes, thus better exploiting the graph structural information in the graph pooling. Moreover, the proposed LiftPool is guaranteed to be localized and only introduce small parameter and computational complexity. By combining the proposed graph lifting structures with permutation-invariant node selection methods (e.g, SAGPool \cite{lee2019self}), the resulting graph pooling operations are also guaranteed to be permutation-invariant. We evaluate the proposed LiftPool in graph classification tasks on a collections of benchmark datasets, demonstrating a significant performance gain over existing state-of-the-art graph pooling methods. 

\section{Related Work}
\textbf {Graph Convolution.} Graph convolution can be roughly grouped into two categories: spatial-based and spectral-based approaches. Spatial-based approaches \cite{hamilton2017inductive,velivckovic2017graph,wang2019dynamic, xu2018powerful} directly generalize classical slide-window based convolution scheme to graph domains, where central node aggregates features from its neighbors. On the other hand, spectral-based \cite{bruna2013spectral,defferrard2016convolutional,henaff2015deep,kipf2016semi} graph convolutions are defined in graph Fourier domain via convolution theorem by filtering graph signals with spectral graph filters. Most of these works fall into the message passing framework \cite{gilmer2017neural}, where node representations are learned by aggregating information from adjacent nodes. Advanced methods such as attention mechanism \cite{velivckovic2017graph} and more complicated spectral graph filters \cite{bianchi2019graph,levie2018cayleynets} are widely studied to improve the model capacity, resulting in state-of-the-art performance in various graph learning tasks. 

\textbf {Graph Pooling.} Generalizing pooling to irregular graphs is challenging. Previous works for graph-level representation learning (e.g., graph classification) usually adopt global pooling methods to summarize node features, where aggregation functions or neural networks are adopted to summarize all the node representations. For example, Set2Set \cite{vinyals2015order} adopts a learnable LSTM to aggregate information from all nodes, and in SortPool \cite{zhang2018end}, nodes are sorted in a descending order according to their structural roles. However, global pooling methods cannot learn hierarchical feature representations, and thus fail to capture graph structure information well.

Hierarchical pooling methods also fall into two classes: clustering-based and downsampling-based methods. Earlier clustering-based pooling methods adopt graph coarsening algorithms, such as spectral clustering algorithm \cite{karypis1998fast, von2007tutorial} and Graclus methods \cite{defferrard2016convolutional, dhillon2007weighted}, which are however very time consuming and only suitable for preprocessing fixed graph structures. Later, neural network-based methods are proposed. For example, DiffPool \cite{ying2018hierarchical} use an additional graph neural network to softly assign nodes to different clusters to compute dense cluster assignment matrix for graph coarsening, whose complexity is quadratic to graph size and is thus non-scalable. Furthermore, coarsened graphs constructed with clustering centroids would significantly destroy the original graph structures. In contrast, downsampling-based methods are much more efficient and able to well preserve the original graph structures, thus having attracted great interests. For example, gPool~\cite{gao2019graph} preserves the top-$k$ nodes according to scores obtained by projecting the node features into a learnable feature vector. SAGPool~\cite{lee2019self} adopts self-attention mechanism to consider both graph topology and node feature information for most informative nodes, achieving the state-of-the-art performance. However, both of them follow a two-stage strategy, thus inherently suffering from the information loss problem.

\section{The Proposed Method}
\label{Me}
  
We begin with first recalling classical lifting structures for signal compression in Sec.~\ref{sec31}. Then, we introduce the framework of hierarchical graph neural networks and different graph pooling strategy in Sec.~\ref{sec32}. Finally, we detail the proposed LiftPool and the graph lifting structures in Sec.~\ref{sec33}.
 
 \subsection{Lifting Structure for Signal Compression} \label{sec31}
\begin{figure*}
  \centering
  \subfigure{\includegraphics[width=1\columnwidth]{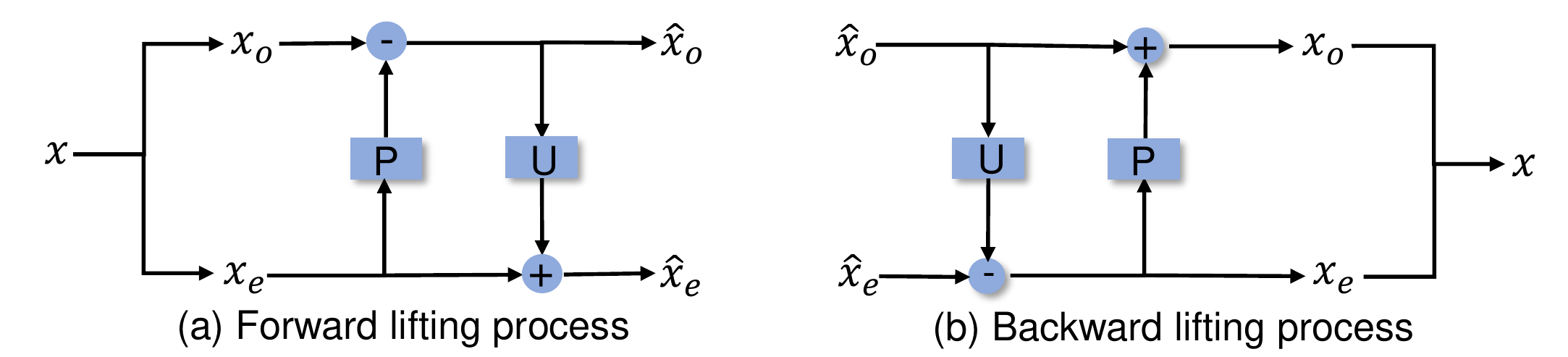}} 
  \subfigure{\includegraphics[width=\columnwidth]{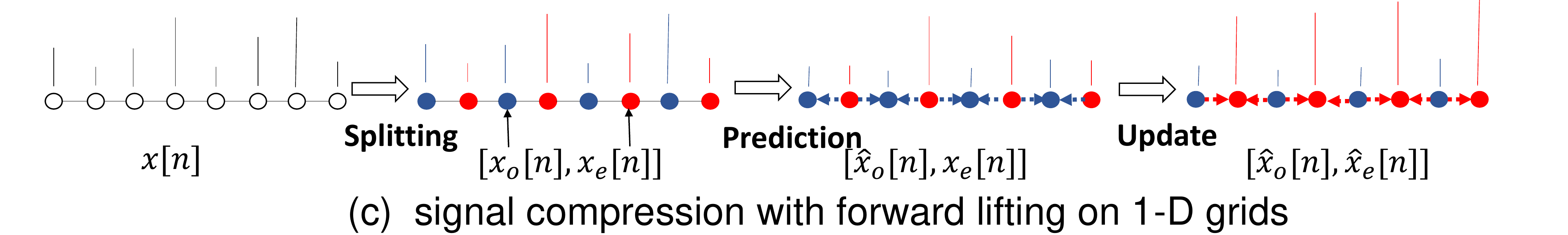}}  \caption{Illustration of entire lifting structures which consists of forward and backward lifting. The forward lifting is composed of three operations: signal splitting, prediction and update. The bottom is a simple illustration for lifting in signal compression.}
  \label{fig2}
\end{figure*}

Different from classical spectral-based transforms, lifting structure \cite{sweldens1998lifting} is characterized by the entire spatial implementation, providing us with an easy control of locality and sparse signal subset in spatial domain. As illustrated in Fig.~\ref{fig2}, an entire lifting structure is composed of two processes: a forward lifting and a backward lifting process, which performs forward and inverse transform, respectively. The forward lifting process consists of three main operations: signal splitting, prediction and update. Here, for simplicity, we describe the lifting structure over a discrete signal $x[n]$ residing on 1-D grids as shown in Fig.~\ref{fig2}(c). Specifically, the signal $x[n]$ is firstly split into two disjoint subsets, an odd subset $x_o=x[2n+1]$ and an even subset $x_e=x[2n]$. Note that we take an equal-splitting here for example, and in fact, any non-overlaping partition of $x[n]$ is also possible, which provides us with the flexibility to partition the signal into two arbitrary disjoint subsets. Then, a prediction operation is adopted to obtain the high-frequency signal presentation $\hat{x}_o$ (local information) by subtracting low-frequency representation predicted from even signals ${x}_e$ with a prediction operator $P$, given by 
\begin{equation}\label{eq1}
\hat{x}_o=x_o-Px_e.
\end{equation} 
Next, an update operation is adopted to process these high-frequency signals (local information) and propagate them to $x_e$ with an update operator $U$, which can be formulated as
\begin{equation}\label{eq2}
\hat{x}_e=x_e+U\hat{x}_o.
\end{equation}
Through lifting, $x$ is compressed on the even subset, obtaining a coarser approximation $\hat{x}_e$. Essentially, the lifting process is a decorrelation between $x_e$ and $x_o$, where the redundant global information of $x_o$ that can be predicted with $x_e$ is removed while the local information that is distinct on $x_o$ is distilled and propagated to $x_e$ to generate a more accurate approximation $\hat{x}_e$. This process can be simplified as Fig.\ref{fig2}(c). 
Inspired by the signals compression, in this paper, we propose to distill and maximally preserve the local structural information with an additional forward graph lifting structure, to address the information loss problem in existing graph pooling methods.

\subsection{Framework of Hierarchical Graph Representation Learning} \label{sec32}

{\bf Notations:} Consider an undirected graph ${\mathcal G}=({\mathcal V},{\mathcal E}, W)$, where $\mathcal V$ and $\mathcal E$ are the set of nodes and set of edges, respectively. The adjacency matrix $W$ represents the graph topology with its entry $W_{ij}$ for the edge connecting node $i$ and $j$ and the degree matrix $D$ is a diagonal matrix with $D_{ii}=\sum_jW_{ij}$. Let us denote $W_n=D^{-1}WD^{-1}$ the normalized adjacency matrix derived from $A$ and $D$ and $ W_a=W_n+\lambda I$ its augmented version with added self-loop. We introduce the subscript $l$ to indicate the $l$-th layer of GNNs, where $X_l=[x_{l,1},x_{l,2}, \cdots, x_{l,|\mathcal{V}_l|}]\in\mathbb{R}^{|\mathcal{V}_l|\times d_{l-1}}$ represents the $d_{l-1}$ dimensional node features residing on the input graph $\mathcal{G}_{l}=(\mathcal{V}_{l},\mathcal{E}_{l},A_l)$ with $|\mathcal{V}_l|$ nodes. Without loss of generality, we consider the $l$-th layer in the remaining of this section.

\textbf{Graph Convolution Layer:} Most existing graph convolutions follow the message passing scheme \cite{gilmer2017neural}. The output graph feature map $H_{l}=\sigma(m(G_l, X_l, \Theta_l))\in\mathbb{R}^{|\mathcal{V}_l|\times d_{l}}$ is obtained from $X_l$ using the structure-aware graph shift matrix $G_l$ (\emph{e.g.}, $W_n$ or $W_a$) and the information aggregation function $m()$ with trainable parameters $\Theta_l$ and the nonlinear activation function $\sigma()$.
The graph shift matrix and information aggregation function vary for different variants of graph convolutions. In this paper, we formulate $H_{l}=ReLU({W}_{la}X_l\Theta_l)$ as the widely used GCN~\cite{kipf2016semi}, where the linear combination function, augmented graph adjacency matrix $W_{la}$ and $ReLU$ activation function are adopted. Note that LiftPool is general to support different graph convolutions, \emph{e.g}, GraphSAGE~\cite{hamilton2017inductive} and GAT~\cite{velivckovic2017graph}.

\textbf{Two-Stage Graph Pooling:} When $H_{l}$ on ${\mathcal G}_l=({\mathcal V}_l,{\mathcal E}_l,W_l)$ is extracted, it is fed into the graph pooling operation to produce the coarsened graph representation $X_{l+1}$ on   ${\mathcal G}_{l+1}=({\mathcal V}_{l+1},{\mathcal E}_{l+1},W_{l+1})$. Existing pooling methods~\cite{gao2019graph,lee2019self} commonly consist of two stages as below.

\textbf{Stage I (Node Selection):} Information or structure-based criteria are commonly utilized to select the most informative nodes. Specifically, nodes are sorted with the assigned importance scores $S_l=f_l(H_l)\in {\mathbb R}^{|{\mathcal V_l}|}$ computed from $H_l$ using the employed node selection function $f_l$.
The top-ranked $|{\mathcal V}_{l+1}|=|{\mathcal V}^p_{l}|$ nodes are to be preserved, while the rest $|{\mathcal V}_l^r|$=$|{\mathcal V}_l|-|{\mathcal V}_l^p|$ nodes will be removed in graph coarsening. Here, we use ${\mathcal V}_l^p$ and ${\mathcal V}_l^r$ to represent  the subset of nodes to be preserved and the subset of nodes to be removed, respectively. 

\textbf{Stage II (Graph Coarsening):} The graph coarsening matrix $D_{l}\in {\mathbb R}^{|{\mathcal V}_{l+1}|\times |{\mathcal V}_{l} |}$ is obtained according to ${\mathcal V}_l^p$ and ${\mathcal V}_l^r$. The $(i,j)$-th entry $D_{l}(i,j)$ of $D_l$ is calculated by
\begin{equation}\label{eq3}
D_{l}(i,j)=\left\{
\begin{aligned}
1,&\quad j={\mathcal V}_{l+1}(i)\\
0,&\quad \text{otherwise}.
\end{aligned}
\right.
\end{equation}
According to Eq.~\eqref{eq3}, there is only one non-zero value for each row in $D_l$. The coarsened graph features $X_{l+1}\in {\mathbb R}^{|{\mathcal V}_{l+1}|\times d_l}$ and graph adjacent matrix $W_{l+1}$ are obtained using $D_l$ as
\begin{align}\label{eq4}
X_{l+1}=&D_lH_{l}\nonumber,\\
W_{l+1}=&D_l^TW_lD_l.
\end{align}
Since $\mathcal{V}_l$ and $X_l$ are inherently coupled, the features of nodes in ${\mathcal V}_l^r$ would also be lost. This fact suggests significant information loss, as these features encode their local structural information.
 
{\bf LiftPool:} To solve the inherent limitation of two-stage pooling strategy, LiftPool introduces an additional graph lifting stage to propagate local structural information from ${\mathcal V}_l^r$ to ${\mathcal V}_l^p$. Thus, graph coarsening would drop less structural information and yield more effective hierarchical graph representation. In the proposed three-stage strategy, the additional stage of graph lifting is inserted between the stages of node selection and graph coarsening. When ${\mathcal V}_l^p$ and ${\mathcal V}_l^r$ are obtained by node selection, graph lifting distills the local structural information of ${\mathcal V}_l^r$ and propagates it to ${\mathcal V}_l^p$ with a lifting structure. The graph feature maps $H_l^p \in {\mathbb R}^{|{\mathcal V}_l^p|\times d_l}$ on ${\mathcal V}_l^p$ and $H_l^r \in {\mathbb R}^{|{\mathcal V}_l^r|\times d_l}$ on ${\mathcal V}_l^r$ are transformed to generate the lifted graph feature maps $\hat{H}_l^p \in {\mathbb R}^{|{\mathcal V}_l^p|\times d_l}$ on ${\mathcal V}_l^p$ and $\hat{H}_l^r \in {\mathbb R}^{|{\mathcal V}_l^r|\times d_l}$ on ${\mathcal V}_l^r$ as
\begin{equation}\label{eq5}
[\hat{H}_l^p,\hat{H}_l^r]=L_{\Theta'_l}(H_l^p,H_l^r),
\end{equation}
where $L_{\Theta'_l}$ is the graph lifting structure parametrized with ${\Theta'_l}$. Consequently, the enhanced nodes features $\hat{H}_l^p$ on ${\mathcal V}_l^p$ are used to calculate $W_{l+1}$ and improve the coarsened graph representation $X_{l+1}$.

\subsection{Lifting-based Graph Pooling}\label{sec33}
In this subsection, we elaborate the three stages, \emph{i.e.}, \emph{permutation-invariant node selection}, \emph{graph lifting} and \emph{graph coarsening}, of the proposed LiftPool.

\textbf{Permutation-invariant Node Selection:} We adopt the permutation-invariant attention-based methods in SAGPool~\cite{lee2019self} for node selection. The self-attention mechanism is leveraged to jointly consider graph topology and node features in node selection. An additional GCN layer is developed to yield the attention scores $S_{l}$ to determine the importance of each node. 
\begin{equation}\label{eq6}
S_{l}=\sigma({W_a}_l H_l \Theta^s_l),
\end{equation}
where $\sigma$ is the activation function (e.g., \emph{tanh}), ${W_a}_l $ is the augmented normalized adjacency matrix and $\Theta^s_l$ is the parameters. Given the predefined pooling ratio $\eta$, ${\mathcal V}^p_{l}$ and ${\mathcal V}^r_{l}$ are determined according to $S_{l}$ with $|{\mathcal V}^p_{l}|=\eta\cdot|{\mathcal V}_l|$ and $|{\mathcal V}^r_{l}|=|{\mathcal V}_l|-|{\mathcal V}^p_{l}|$. 
In Section~\ref{sec4}, we show that, when node selection is permutation-invariant, permutation invariance is also guaranteed for the proposed LiftPool. 

\textbf{Graph Lifting:} Before graph coarsening, we distill the local structural information of $\mathcal{V}^r_l$ and propagate it to ${\mathcal V}_{l}^p$ via a graph lifting structure. Similar to classical lifting scheme, the proposed graph lifting structure also consists of three operations, \emph{i.e.}, splitting, prediction and update. First, ${H_l}$ can be naturally split into two disjoint subsets $H_l^p$ and $H_l^r$ that reside on ${\mathcal V}^p_{l}$ and ${\mathcal V}^r_{l}$, respectively. Subsequently, $H_l^p$ is used to predict the global information with the prediction operator $P_{\Theta_l^P}$ and the local structural information of ${\mathcal V}^r_{l}$ is obtained by subtracting this global information. Finally, the update operator $U_{\Theta_l^U}$ is learned to align and propagate the local information to ${\mathcal V}^p_{l}$. Thus, local structural information of ${\mathcal V}^r_{l}$ is transformed to ${\mathcal V}^p_{l}$ for graph coarsening. We formulate this stage as
\begin{align}\label{eq7}
\hat{H}_l^r&=H_l^r-P_{\Theta^P_l}(H_l^p),\nonumber\\
\hat{H}_l^p&=H_l^p+U_{\Theta^U_l}(\hat{H}_l^r),
\end{align}

where ${\Theta^P_l}$ and ${\Theta^U_l}$ is the learnable parameters for the prediction and update operator, respectively.

In hierarchical graph representation learning, spatially localized and permutation-invariant operations are usually preferred. Furthermore, the extra computational complexity and parameters should also be reasonable for fast inference and suppressing overfitting. Therefore, GCN-like graph lifting operations are developed to satisfy these requirements.  
\begin{align}\label{eq8}
P_{\Theta^P_l}({H}_l^p)&=ReLU({W_l}_a^{pr}{H}_l^p\Theta_l^P),\nonumber\\
U_{\Theta^U_l}(\hat{H}_l^r)&=ReLU({W_l}_a^{rp}\hat{H}_l^r\Theta_l^U).
\end{align}
Here, ReLU activation function is adopted to enhance the model capacity with nonlinear lifting. ${W_l}_a^{pr}$ is the submatrix of ${W_l}_a$ that represents the edges connecting ${\mathcal V}^p_{l}$ to ${\mathcal V}^r_{l}$, while ${W_l}_a^{rp}$ represents the edges connecting ${\mathcal V}^r_{l}$ to ${\mathcal V}^p_{l}$. Note that we have ${W_l}_a^{pr}={W_l}_a^{rp}$ for undirected graphs. $\Theta_l^p=\text{diag}(\theta_l^{p1},\cdots,\theta_l^{pd_l})\in {\mathbb R}^{d_l\times d_l}$ and $\Theta_l^u=\text{diag}(\theta_l^{u1},\cdots,\theta_l^{ud_l}) \in {\mathbb R}^{d_l\times d_l}$ are the parameters that perform as scale factors to control the information propagation in each feature channel. It is worth mentioning that multiple lifting layers can be stacked for more powerful models. 

{\bf Graph Coarsening:} Graph lifting compensates the preserved ${\mathcal V}^p_{l}$ with the transformed local structural information of ${\mathcal V}^r_{l}$. Finally, we calculate the reduced graph with the coarsening matrix $D_l$ and the lifted graph representation $\hat{H}_l=[\hat{H}_l^p,\hat{H}_l^r]$ according to Eq.~\eqref{eq3} and Eq.~\eqref{eq4}. 

\section{Properties of LiftPool}\label{sec4}
\begin{figure}[!t]
  \centering
  \includegraphics[width=0.7\columnwidth]{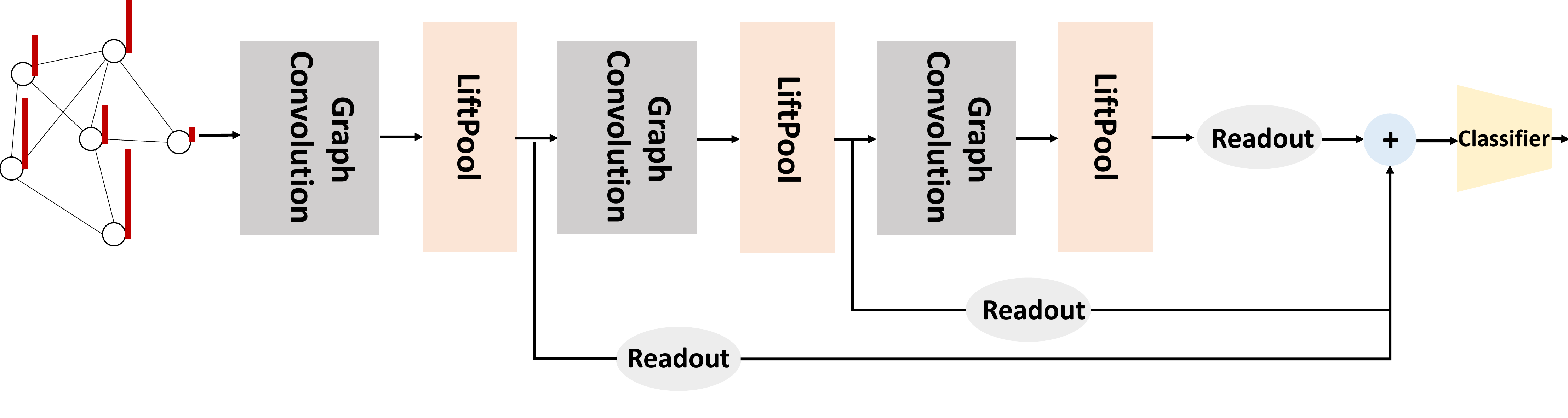}
  \caption{Overall structures of the adopted hierarchical graph neural network.}
  \label{fig3}
  \end{figure}

This section first presents the desirable properties of LiftPool such as locality and permutation invariance, and subsequently discusses the computational and parametric complexity.
\begin{pro}\label{pro1}
Given one graph lifting layer consisting of one prediction and one update operation shown in Eq.~\eqref{eq7} and \eqref{eq8}, LiftPool is localized within $2$-hops in the spatial domain.
\begin{proof} Please refer to Section~2 in the Appendix.
\end{proof}
\end{pro}
It is also necessary for GNNs to be invariant to node permutation for isomorphic graphs. LiftPool is guaranteed to be permutation-invariant with a mild requirement on the node selection methods.

\begin{pro}\label{pro2}
If the importance score function $f$ is invariant to node permutation, LiftPool is guaranteed to be permutation invariant.
\begin{proof}
Please refer to Section~2 in the Appendix.
\end{proof}
\end{pro}
 
{\bf Computational and Parametric Complexity:} LiftPool requires an additional stage of graph lifting stage in comparison to existing pooling methods, \emph{e.g}, SAGPool. Here, we make a discussion on its computational and parametric complexity. Given $H_l\in{\mathbb R}^{|{\mathcal V}_l|\times d_l}$, LiftPool requires only $2d_l$ additional parameters in ${\Theta^P_l}$ and ${\Theta^U_l}$ with a moderate number of channels (\emph{e.g.}, $d_l$ is 64 or 128). Note that the extra parametric complexity is independent of the graph size. The maximum computation complexity introduced by the graph lifting operations is ${\mathcal O}(d_l|\mathcal E|)$ only when the graph is bipartite by the preserved and removed nodes. Here, $\mathcal |\mathcal E|$ is the number of edges in the graph, which is small and decreases rapidly with the growth of the number of pooling layers. Therefore, the additional computational and parametric complexity introduced by LiftPool is reasonable.

\section{Experiments}
  \begin{table}
  \caption{Experimental results with 20 random seeds. We compare global pooling methods ($_g$) with hierarchical pooling methods ($_h$).}
  \label{tab2}
  \centering
  \resizebox{0.9\textwidth}{!}{
  \begin{tabular}{lccccc}
    \toprule               
      Models &D$\&$D&PROTEINS& NCI1 &NCI109&FRANKEN  \\
      \midrule
   Set2Set$_g^\sharp$  &71.60 $\pm$ 0.87  &72.16 $\pm$ 0.43    &66.97 $\pm$ 0.74 &61.04 $\pm$ 2.69 &61.46 $\pm$ 0.47\\
   Sortpool$_g^\sharp$  &71.87 $\pm$ 0.96  &73.91 $\pm$ 0.72  &68.74 $\pm$ 1.07 &68.59 $\pm$ 0.67 &63.44 $\pm$ 0.65 \\
    \midrule
    DiffPool$_h^\star$    &66.95 $\pm$ 2.41 &68.20 $\pm$ 2.02 &62.32 $\pm$ 1.90 &61.98 $\pm$ 1.98& 60.60 $\pm$ 1.62 \\
    gPool$_h^\star$     &75.01 $\pm$ 0.86 & 71.10 $\pm$ 0.90 & 67.02 $\pm$ 2.25& 66.12 $\pm$ 1.60 &61.46 $\pm$ 0.84\\
    SAGPool$_h^\star$&76.45 $\pm$ 0.97& 71.86 $\pm$ 0.97 &67.45 $\pm$ 1.11 &67.86 $\pm$ 1.41&61.73 $\pm$ 0.76 \\
    \midrule
    LiftPool$_h$&\bf 76.61 $\pm$ 1.12&\bf 74.09 $\pm$ 0.85& \bf 71.82 $\pm$ 1.18& \bf 71.39 $\pm$ 0.86& 61.34 $\pm$ 1.46\\
    \bottomrule
  \end{tabular}}
\end{table}
\label{Ex}

\subsection{ Experimental Settings}
{\bf Datasets and baselines:}
We select five graph classification benchmark datasets, the same as in SAGPool and with different graph sizes. Detailed statistics are presented in Section 1 of the Appendix. Specifically, D$\&$D consists of large protein graphs
that are classified into enzyme or non-enzyme. PROTEINS is also a protein dataset with medium-size graphs where the nodes represent elements with secondary structures. 
NCI1 and NCI109 are two medium-size biological datasets for classifying activity against non-small cell lung cancer and ovarian cancer cell lines, with each graph representing a chemical compound. 
FRANKENSTEIN is a small molecular graph dataset for classifying whether a molecular is a mutagen or non-mutagen. We compare our methods with the state-of-the-art pooling methods including two global pooling methods: Set2Set \cite{vinyals2015order} and SortPool \cite{zhang2018end} and three hierarchical pooling methods: DiffPool \cite{ying2018hierarchical}, gPool \cite{gao2019graph}, SAGPool \cite{lee2019self}. 
We adopt the results reported in \cite{lee2019self, ranjan2019asap} for a fair comparison in Table~\ref{tab2}, marked with upscripts $\star$ and $\sharp$, respectively.

{\bf Model architectures and training protocol:} 
Fig.~\ref{fig3} illustrates the model architectures. Each feature extraction layer consists of a graph convolution layer and a pooling layer and three layers are stacked for hierarchical features learning. 
We adopt the same training procedure as in SAGPool. 10-fold cross validation with 20 random initiation for each dataset are utilized for evaluation. Table~\ref{tab2} reports the average test accuracy with standard deviation for the total 200 testing results.Note that due to the restriction of storage, the batch size of our model for DD is reduced to 28. Please refer to Section 1 in the Appendix for more details about the model configurations as well as hyper-parameters. 

\subsection{Results and Analysis}

\begin {figure}
  \centering
  \includegraphics[width=0.75\columnwidth,height=0.3\columnwidth]{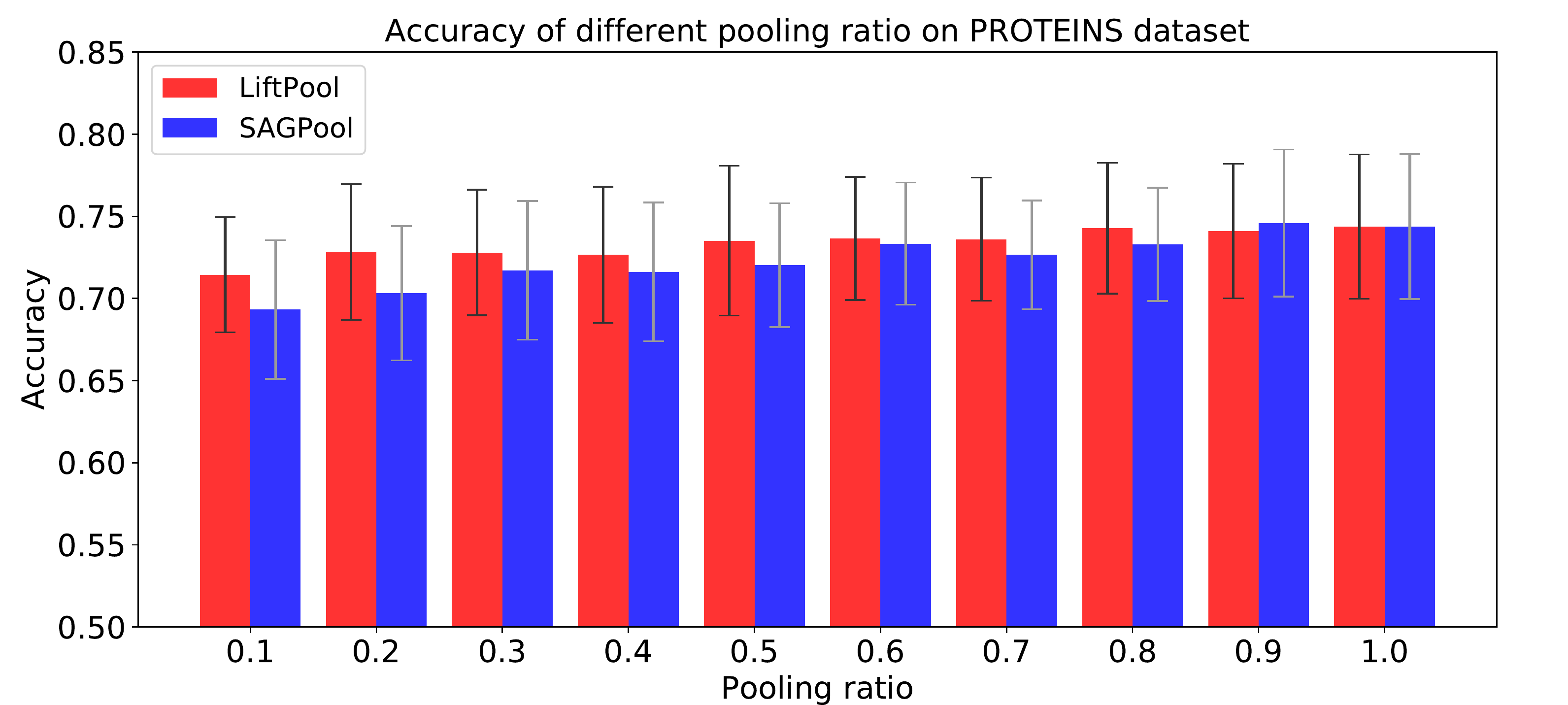}
  \caption{Comparison between LiftPool and SAGPool on PROTEINS dataset with different pooling ratio}
  \label{fig4}
  \end{figure}
  
  \begin{table}
  \caption{Performance with variants of graph convolution on DD and PROTEINS datasets}
  \centering
  \resizebox{0.7\textwidth}{!}{
  \begin{tabular}{lcccc}
    \toprule               
    Datasets &Model  & GCN   &GAT & GraphSAGE   \\
    \midrule
    \multirow{2}*{D$\&$D}  &SAGPool$^\star$&  76.45 $\pm$ 0.97  & 75.49 $\pm$ 0.93 & 76.28 $\pm$ 1.06\\
    &LiftPool & \bf 76.61 $\pm$ 1.12  & 76.15 $\pm$ 0.95 & 76.21 $\pm$ 0.95  \\
    \midrule
    \multirow{2}*{PROTEINS}& SAGPool$^\star$& 71.86 $\pm$ 0.97 &  71.98 $\pm$ 1.01 & 71.93 $\pm$ 0.82 \\
    & LiftPool & 74.09 $\pm$ 0.85 & 74.34 $\pm$ 0.78 & \bf 74.39 $\pm$ 0.51    \\
    \bottomrule
  \end{tabular}}
  \label{tab3}
\end{table}

{\bf Hierarchical vs Global Pooling:} Table~\ref{tab2} presents the performance of global and hierarchical pooling methods on datasets containing graphs of different sizes. With regard to large graphs (DD), hierarchical pooling methods significantly outperform global ones  as they can better exploit the multi-scale structures of large graphs. For medium-size (PROTEINS, NCI1 and NCI109) and small (FRANKENSTEIN) graphs, 
previous graph pooling methods (SAGPool and gPool) tend to lose excessive structural information, which thus lead to similar or even inferior performance in comparison to global pooling methods. In contrast, our methods is able to preserve more local structural information with the additional graph lifting stage. Consequently, we still substantially surpass global pooling methods, which demonstrates the effectiveness of our methods.

{\bf Comparison of Hierarchical Methods:}
SAGPool surpasses gPool on all datasets, since graph topology and node features are jointly considered to select the most informative nodes. We improve SAGPool by maximally preserving and utilizing the local structural information of the removed nodes, and thus consistently outperform all other hierarchical pooling methods on most of the datasets, especially on PROTEINS, NCI1 and NCI109 datasets that consist of medium-sized and highly-irregular graphs, with the gains of $2.23\%$, $4.37\%$ and $3.53\%$, respectively. With regard to DD dataset, less local information can be exploited since the graphs are more regular (i.e., the degree distribution tends to be uniform), which may lead to smooth graph features with message passing graph convolutions. For FRANKENSTEIN, as the graphs are very small and sparse, the inferior performance is resulted from its sparse connections between the preserved and removed nodes, which hinders the information propagation in graph lifting. To verify this, we further strengthen the information propagation by connecting nodes that are reachable within two and three hops, and consequently, the performance is then increased to $61.71 \pm 1.01$ and $62.20 \pm 0.97$, respectively.

\subsection{Ablation Study}
{\bf Pooling Ratios:} To further demonstrate the superiority and stability of LiftPool, we compare LiftPool and SAGPool in different pooling ratios. For each pooling ratio, we train and evaluate both models with 20 random seeds and the mean results together with their standard deviation ( error bar ) are presented in Fig.~\ref{fig4}. It can be observed that LiftPool consistently outperforms SAGPool in small pooling ratio (less than 0.9), as more structural information can be better exploited with the additional graph lifting stage. Furthermore, for both models, the performance is quickly improved in small pooling ratio (0.1-0.5) while fluctuated in large pooling ratios (0.5-1.0). This phenomenon can be explained as graphs containing redundant information. In small pooling ratios, increasing the pooling ratio helps to preserve more effective information, while in large pooling ratios, more redundant information is introduced, which does no help for, sometimes may even degrade the performance.  

{\bf Variants of GNNs for Node Selection:} We also adopt variants of graph convolution to calculate importance scores for LiftPool and compare them with SAGPool. The performance of different models on DD and PROTEINS datasets are presented in Table~\ref{tab3}. It can be seen that our methods consistently outperforms SAGPool across different graph convolutions, which empirically shows the effectiveness and stability of LiftPool. Note that our best results on DD and PROTEINS dataset are achieved with GCN and GraphSAGE, respectively, reveling that there may be different graph convolutions that are suitable for different graphs. 
More experimental results including comparison of different lifting schemes and illustration of pooled graphs. are presented in the Appendix.

\section{Conclusion}
In this paper, we proposed a novel three-stage strategy for graph pooling to improve the hierarchical graph representation. A lifting-based graph pooling, named LiftPool, was developed to maximally preserve the local structural information of graphs where an additional graph lifting stage was introduced. Experimental results on benchmark graph classification datasets have demonstrated the effectiveness of the LiftPool for effective graph pooling. As lifting structure has been shown to be suitable for multiscale signal processing where information can be perfectly recovered, in the future, we will develop a general multi-scale framework via lifting structure for graph signal processing, and theoretical properties such as stability and capacity will also be explored.

\section{Border Impact}
As a large number of data can be represented as graphs, e.g, social networks, protein networks and chemical networks. The proposed graph pooling method that facilitates efficient and effective hierarchical graph representation learning  will help researcher in these areas, especially for drug discovery and community detection. It can also help us to acquire better understanding of proteins, molecules as well as chemical compounds by analyzing their structures. 

\bibliographystyle{plain}
\bibliography{reference}

\begin{thebibliography}{10}

\bibitem{bahdanau2014neural}
Dzmitry Bahdanau, Kyunghyun Cho, and Yoshua Bengio.
\newblock Neural machine translation by jointly learning to align and
  translate.
\newblock In {\em 3rd International Conference on Learning Representations},
  2015.

\bibitem{bianchi2019graph}
Filippo~Maria Bianchi, Daniele Grattarola, Cesare Alippi, and Lorenzo Livi.
\newblock Graph neural networks with convolutional {ARMA} filters.
\newblock {\em arXiv preprint arXiv:1901.01343}, 2019.

\bibitem{bruna2013spectral}
Joan Bruna, Wojciech Zaremba, Arthur Szlam, and Yann LeCun.
\newblock Spectral networks and locally connected networks on graphs.
\newblock {\em arXiv preprint arXiv:1312.6203}, 2013.

\bibitem{davidson2002genomic}
Eric~H Davidson, Jonathan~P Rast, Paola Oliveri, Andrew Ransick, Cristina
  Calestani, Chiou-Hwa Yuh, Takuya Minokawa, Gabriele Amore, Veronica Hinman,
  Cesar Arenas-Mena, Ochan Otim, C.~Titus Brown, Carolina~B. Livi, Pei~Yun Lee,
  Roger Revilla, Alistair~G. Rust, Zheng~Jun Pan, Maria~J. Schilstra, Peter
  J.~C. Clarke, Maria~I. Arnone, Lee Rowen, R.~Andrew Cameron, David~R. McClay,
  Leroy Hood, and Hamid Bolouri.
\newblock A genomic regulatory network for development.
\newblock {\em Science}, 295(5560):1669--1678, 2002.

\bibitem{defferrard2016convolutional}
Micha{\"e}l Defferrard, Xavier Bresson, and Pierre Vandergheynst.
\newblock Convolutional neural networks on graphs with fast localized spectral
  filtering.
\newblock In {\em Advances in Neural Information Processing Systems 29}, pages
  3844--3852, 2016.

\bibitem{dhillon2007weighted}
Inderjit~S. Dhillon, Yuqiang Guan, and Brian Kulis.
\newblock Weighted graph cuts without eigenvectors: A multilevel approach.
\newblock {\em IEEE Transactions on Pattern Analysis and Machine Intelligence},
  29(11):1944--1957, 2007.

\bibitem{duvenaud2015convolutional}
David~K. Duvenaud, Dougal Maclaurin, Jorge Iparraguirre, Rafael Bombarell,
  Timothy Hirzel, Al{\'a}n Aspuru-Guzik, and Ryan~P. Adams.
\newblock Convolutional networks on graphs for learning molecular fingerprints.
\newblock In {\em Advances in Neural Information Processing Systems 28}, pages
  2224--2232, 2015.

\bibitem{gao2019graph}
Hongyang Gao and Shuiwang Ji.
\newblock Graph {U-Nets}.
\newblock In {\em Proceedings of the 36th International Conference on Machine
  Learning}, pages 2083--2092, 2019.

\bibitem{gilmer2017neural}
Justin Gilmer, Samuel~S. Schoenholz, Patrick~F. Riley, Oriol Vinyals, and
  George~E. Dahl.
\newblock Neural message passing for quantum chemistry.
\newblock In {\em Proceedings of the 34th International Conference on Machine
  Learning}, pages 1263--1272, 2017.

\bibitem{hamilton2017inductive}
Will Hamilton, Zhitao Ying, and Jure Leskovec.
\newblock Inductive representation learning on large graphs.
\newblock In {\em Advances in Neural Information Processing Systems 30}, pages
  1024--1034, 2017.

\bibitem{he2016deep}
Kaiming He, Xiangyu Zhang, Shaoqing Ren, and Jian Sun.
\newblock Deep residual learning for image recognition.
\newblock In {\em 2016 IEEE Conference on Computer Vision and Pattern
  Recognition (CVPR)}, pages 770--778, 2016.

\bibitem{henaff2015deep}
Mikael Henaff, Joan Bruna, and Yann LeCun.
\newblock Deep convolutional networks on graph-structured data.
\newblock {\em arXiv preprint arXiv:1506.05163}, 2015.

\bibitem{hinton2012deep}
Geoffrey Hinton, Li~Deng, Dong Yu, George~E. Dahl, Abdel-rahman Mohamed,
  Navdeep Jaitly, Andrew Senior, Vincent Vanhoucke, Patrick Nguyen, Tara~N.
  Sainath, and Brian Kingsbury.
\newblock Deep neural networks for acoustic modeling in speech recognition: The
  shared views of four research groups.
\newblock {\em IEEE Signal Processing Magazine}, 29(6):82--97, 2012.

\bibitem{karpathy2014large}
Andrej Karpathy, George Toderici, Sanketh Shetty, Thomas Leung, Rahul
  Sukthankar, and Li~Fei-Fei.
\newblock Large-scale video classification with convolutional neural networks.
\newblock In {\em 2014 IEEE Conference on Computer Vision and Pattern
  Recognition}, pages 1725--1732, 2014.

\bibitem{karypis1998fast}
George Karypis and Vipin Kumar.
\newblock A fast and high quality multilevel scheme for partitioning irregular
  graphs.
\newblock {\em SIAM Journal on Scientific Computing}, 20(1):359--392, 1998.

\bibitem{kipf2016semi}
Thomas~N. Kipf and Max Welling.
\newblock Semi-supervised classification with graph convolutional networks.
\newblock In {\em 5th International Conference on Learning Representations},
  2017.

\bibitem{krizhevsky2012imagenet}
Alex Krizhevsky, Ilya Sutskever, and Geoffrey~E. Hinton.
\newblock {ImageNet} classification with deep convolutional neural networks.
\newblock In {\em Advances in Neural Information Processing Systems 25}, pages
  1097--1105, 2012.

\bibitem{lazer2009social}
David Lazer, Alex Pentland, Lada Adamic, Sinan Aral, Albert-Laszlo Barabasi,
  Devon Brewer, Nicholas Christakis, Noshir Contractor, James Fowler, Myron
  Gutmann, Tony Jebara, Gary King, Michael Macy, Deb Roy, and Marshall
  Van~Alstyne.
\newblock Computational social science.
\newblock {\em Science}, 323(5915):721--723, 2009.

\bibitem{lee2019self}
Junhyun Lee, Inyeop Lee, and Jaewoo Kang.
\newblock Self-attention graph pooling.
\newblock {\em Proceedings of the 36th International Conference on Machine
  Learning}, pages 3734--3743, 2019.

\bibitem{levie2018cayleynets}
Ron Levie, Federico Monti, Xavier Bresson, and Michael~M. Bronstein.
\newblock {CayleyNets}: Graph convolutional neural networks with complex
  rational spectral filters.
\newblock {\em IEEE Transactions on Signal Processing}, 67(1):97--109, 2018.

\bibitem{ranjan2019asap}
Ekagra Ranjan, Soumya Sanyal, and Partha~Pratim Talukdar.
\newblock {ASAP}: Adaptive structure aware pooling for learning hierarchical
  graph representations.
\newblock In {\em Proceedings of the 34th AAAI Conference on Artificial
  Intelligence}, New York, NY, USA, February 2019.

\bibitem{sweldens1998lifting}
Wim Sweldens.
\newblock The lifting scheme: A construction of second generation wavelets.
\newblock {\em SIAM Journal on Mathematical Analysis}, 29(2):511--546, 1998.

\bibitem{velivckovic2017graph}
Petar Veli{\v{c}}kovi{\'c}, Guillem Cucurull, Arantxa Casanova, Adriana Romero,
  Pietro Lio, and Yoshua Bengio.
\newblock Graph attention networks.
\newblock In {\em 6th International Conference on Learning Representations},
  2018.

\bibitem{vinyals2015order}
Oriol Vinyals, Samy Bengio, and Manjunath Kudlur.
\newblock Order matters: Sequence to sequence for sets.
\newblock {\em arXiv preprint arXiv:1511.06391}, 2015.

\bibitem{von2007tutorial}
Ulrike Von~Luxburg.
\newblock A tutorial on spectral clustering.
\newblock {\em Statistics and Computing}, 17(4):395--416, 2007.

\bibitem{wang2019dynamic}
Yue Wang, Yongbin Sun, Ziwei Liu, Sanjay~E. Sarma, Michael~M. Bronstein, and
  Justin~M. Solomon.
\newblock Dynamic graph cnn for learning on point clouds.
\newblock {\em ACM Transactions on Graphics (TOG)}, 38(5):1--12, 2019.

\bibitem{xie2018crystal}
Tian Xie and Jeffrey~C. Grossman.
\newblock Crystal graph convolutional neural networks for an accurate and
  interpretable prediction of material properties.
\newblock {\em Physical Review Letters}, 120(14):145301, 2018.

\bibitem{xu2018powerful}
Keyulu Xu, Weihua Hu, Jure Leskovec, and Stefanie Jegelka.
\newblock How powerful are graph neural networks?
\newblock In {\em 7th International Conference on Leanring Representations},
  2019.

\bibitem{ying2018hierarchical}
Zhitao Ying, Jiaxuan You, Christopher Morris, Xiang Ren, Will Hamilton, and
  Jure Leskovec.
\newblock Hierarchical graph representation learning with differentiable
  pooling.
\newblock In {\em Advances in Neural Information Processing Systems 31}, pages
  4800--4810, 2018.

\bibitem{zhang2018end}
Muhan Zhang, Zhicheng Cui, Marion Neumann, and Yixin Chen.
\newblock An end-to-end deep learning architecture for graph classification.
\newblock In {\em Thirty-Second AAAI Conference on Artificial Intelligence},
  pages 4438--4445, 2018.

\end{thebibliography}
\end{document}